\documentclass[10pt,twocolumn,letterpaper]{article}

\usepackage{iccv}     

\usepackage{graphicx}
\usepackage{amsmath}
\usepackage{amssymb}
\usepackage{booktabs}

\usepackage{amsmath,amsfonts,bm}

\def\eqref#1{equation~\ref{#1}}

\def\1{\bm{1}}

\DeclareMathAlphabet{\mathsfit}{\encodingdefault}{\sfdefault}{m}{sl}
\SetMathAlphabet{\mathsfit}{bold}{\encodingdefault}{\sfdefault}{bx}{n}

\DeclareMathOperator*{\argmax}{arg\,max}
\DeclareMathOperator*{\argmin}{arg\,min}

\usepackage{url}
\usepackage{amsthm}
\usepackage{times}
\usepackage{epsfig}
\usepackage{amsmath}
\usepackage{amssymb}
\usepackage{algorithm}  
\usepackage{algorithmicx}  
\usepackage{algpseudocode} 
\usepackage{booktabs}       
\usepackage{multirow}
\usepackage{graphicx}
\usepackage{wrapfig}
\usepackage{multirow}
\usepackage{float}
\usepackage{mathbbol}

\usepackage[utf8]{inputenc} 
\usepackage[T1]{fontenc}    
\usepackage{url}            
\usepackage{booktabs}       
\usepackage{amsfonts}       
\usepackage{nicefrac}       
\usepackage{microtype}      
\usepackage{xcolor}         

\makeatletter
\@namedef{ver@everyshi.sty}{}
\makeatother
\usepackage{tikz}
\usepackage{bm}
\usepackage{bbm}
\usepackage{paralist}

\newcommand{\tabincell}[2]{\begin{tabular}{@{}#1@{}}#2\end{tabular}}

\newcommand*\circled[1]{\tikz[baseline=(char.base)]{
            \node[shape=circle,draw,inner sep=1pt] (char) {#1};}}

\usepackage[pagebackref=true,breaklinks=true,colorlinks,bookmarks=false]{hyperref}

\usepackage[capitalize]{cleveref}
\crefname{section}{Sec.}{Secs.}
\Crefname{section}{Section}{Sections}
\Crefname{table}{Table}{Tables}
\crefname{table}{Tab.}{Tabs.}

\iccvfinalcopy 

\def\iccvPaperID{9338} 

\ificcvfinal\fi

\begin{document}

\title{Communication-Efficient Vertical Federated Learning\\ with Limited Overlapping Samples}

\author{Jingwei Sun$^1$, Ziyue Xu$^2$, Dong Yang$^2$, Vishwesh Nath$^2$, Wenqi Li$^2$, Can Zhao$^2$, \\
Daguang Xu$^2$, Yiran Chen$^1$, Holger R. Roth$^2$\\
$^1$ Department of Electrical and Computer Engineering, Duke University\\
$^2$ NVIDIA\\
{\tt\small $^1$\{jingwei.sun, yiran.chen\}@duke.edu,}\\
{\tt\small $^2$\{ziyuex, dongy, vnath, wenqil, canz, daguangx, hroth\}@nvidia.com}
}
\maketitle

\ificcvfinal\thispagestyle{empty}\fi

\begin{abstract}
   Federated learning is a popular collaborative learning approach that enables clients to train a global model without sharing their local data. Vertical federated learning (VFL) deals with scenarios in which the data on clients have different feature spaces but share some overlapping samples. Existing VFL approaches suffer from high communication costs and cannot deal efficiently with limited overlapping samples commonly seen in the real world.
   We propose a practical vertical federated learning (VFL) framework called \textbf{one-shot VFL} that can solve the communication bottleneck and the problem of limited overlapping samples simultaneously based on semi-supervised learning. We also propose \textbf{few-shot VFL} to improve the accuracy further with just one more communication round between the server and the clients. In our proposed framework, the clients only need to communicate with the server once or only a few times. We evaluate the proposed VFL framework on both image and tabular datasets. Our methods can improve the accuracy by more than 46.5\% and reduce the communication cost by more than 330$\times$ compared with state-of-the-art VFL methods when evaluated on CIFAR-10. Our code will be made publicly available at \url{https://nvidia.github.io/NVFlare/research/one-shot-vfl}.
\end{abstract}

\section{Introduction}

Federated Learning (FL) is a distributed learning method that enables multiple parties to collaboratively train a model without centralizing their raw data. Therefore, the clients can retain control over their own data assets. FL has received significant attention and has become a major research topic due to its capability to build real-world applications where datasets are isolated across different organizations/devices while preserving data governance and privacy~\cite{li2020federated,kairouz2021advances}.

Existing approaches primarily focus on horizontal federated learning (HFL), where the data from different clients share the same feature space but have different samples~\cite{yang2019federated}. One application of HFL is that smartphones users collaboratively train a next-word prediction model for the smart keyboard~\cite{hard2018federated}. In HFL, the clients are expected to learn the common knowledge from heterogeneous data distributions and produce a global model by aggregating the updates of local models. Hence, the main challenge of HFL is data distribution heterogeneity, and under cross-device scenarios, limited computation resources.

\begin{figure}[t]
\centering
     \includegraphics[scale=0.3]{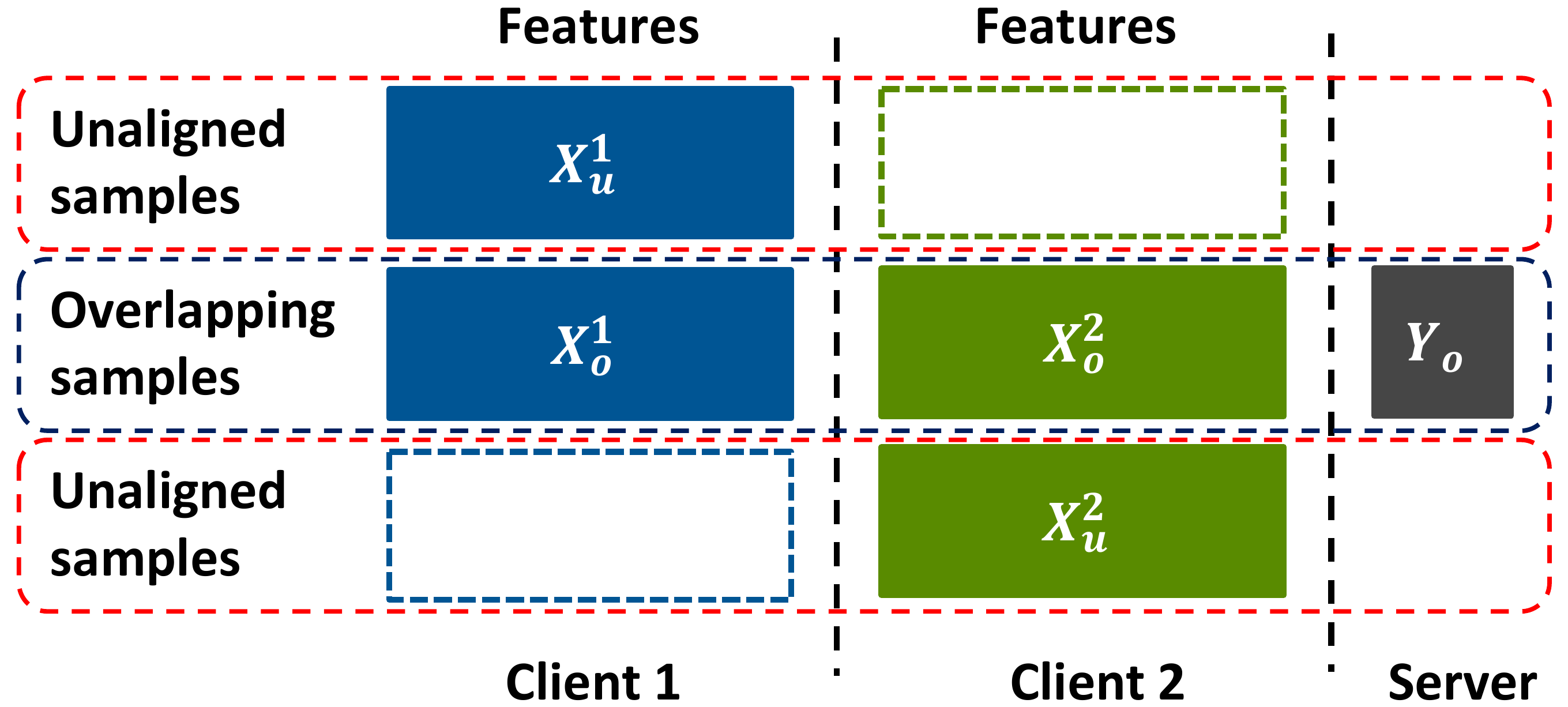}
\caption{An example of data splitting in a two-client VFL setting.}
\label{fig:data_split}
\end{figure}

Vertical federated learning (VFL), on the other hand, focuses on scenarios in which the data on clients have different feature spaces but share some overlapping samples~\cite{yang2019federated}. In addition, the true labels can reside on a third-party server~\cite{romanini2021pyvertical} as shown in \cref{fig:data_split}. For example, a credit bureau collaborates with an e-commerce company and a bank to train a model to estimate a user's credit score. In this case, only the credit bureau has the credit score of the users which will not be shared with the e-commerce company and the bank. VFL is mostly deployed in cross-silo scenarios, and the computation power is usually not a major concern~\cite{kairouz2021advances}. However, VFL faces two unique challenges. First, VFL requires the clients to communicate with the server for each iteration (rather than after several epochs under HFL) of training, which introduces extremely high communication costs. It is also notable that iterative communications require reliable communication channels between the server and the clients, which is usually expensive. In addition to the high communication cost, the other major challenge of VFL is that the number of overlapping samples may be limited. For example, two hospitals in different countries are not expected to have a large number of overlapping patients. The model trained with limited overlapping samples likely cannot achieve reliable performance. 

Furthermore, VFL is currently not as well explored as HFL. Some existing works can reduce the communication cost by reducing the communication frequency or compressing the communicated data~\cite{liu2019communication}. However, most methods only achieve limited reduction from one local update to multiple, while still requiring heavy iterative communications. Other works focus on improving the performance with limited overlapping samples~\cite{kang2022fedcvt, wu2022practical}. Notably, both challenges are bottlenecks of applying VFL in realistic scenarios, and leaving either one unsolved hinders the deployment of VFL in the real world. To the best of our knowledge, there is no work aiming at solving these two challenges simultaneously.

In this paper, we propose \emph{one-shot} VFL, which is a communication-efficient VFL algorithm that can achieve high performance with minimal overlapping samples. In one-shot VFL, the clients are guided to conduct local semi-supervised learning (SSL) using both the overlapping samples and the unaligned samples to train well-performing feature extractors. Under one-shot setting, the clients only need to conduct two upload operations and one download operation for the training session, which drastically reduces the communication cost and frequency. We further propose \emph{few-shot} VFL as an extension of one-shot VFL. Few-shot VFL expands the supervised dataset on clients to improve the performance of the local feature extractors. Compared with one-shot VFL, clients in few-shot VFL conduct one more time of uploading and downloading, but can achieve better performance, especially when the number of overlapping samples is small.

Our key contributions are summarized as follows:
\begin{compactitem}
    \item We propose a communication-efficient VFL algorithm called \emph{one-shot} VFL. To the best of our knowledge, \emph{one-shot} VFL is the first algorithm that can simultaneously address the challenges of high communication cost and limited overlapping samples.
    \item We propose \emph{few-shot} VFL that can improve the performance further under settings with minimal overlapping samples.
    \item We empirically evaluate the performance of \emph{one-shot} VFL and \emph{few-shot} VFL with different data modalities, including image data and tabular data. The results show that our methods improve the accuracy by more than 46.5\% and reduce the communication cost by more than 330$\times$ compared with the state-of-the-art (SOTA) VFL methods on CIFAR-10.
\end{compactitem}
\section{Background and Related Work}

\paragraph{Vertical Federated Learning.}

Vertical federated learning (VFL)~\cite{romanini2021pyvertical,vepakomma2018split} is the concept of collaboratively training a model on a dataset where the clients share some common samples but with different features on each client. VFL was first introduced in \cite{hardy2017private}, where a federated logistic regression algorithm is proposed. 
SecureBoost~\cite{cheng2021secureboost} proposed a secure federated tree-boosting approach in the VFL setting and provided theoretical proof that it achieves the same level of accuracy as its centralized counterparts. Some other gradient boosting tree approaches for VFL include Pivot~\cite{wu2020privacy} and VF$^2$Boost~\cite{fu2021vf2boost}. A federated random forest was also studied in~\cite{liu2020federated}. In addition to tree-based methods, other machine learning algorithms such as linear regression~\cite{zhang2021secure} and logistic regression~\cite{hu2019learning,liu2019communication} have been investigated under VFL settings. However, these algorithms are usually incapable of handling complex tasks such as computer vision (CV) and natural language processing (NLP), in which Deep Neural Networks (DNN) are preferred.
On the neural network side, SplitNN~\cite{vepakomma2018split} was proposed to collaboratively train neural networks by splitting a neural network among participants and exchanging gradients and representations in each iteration. FATE~\cite{liu2021fate} implemented a framework that supports DNN in VFL. 
Even though FATE improves the model's capacity in VFL by supporting DNN, it still requires frequent communication between the participants for each iteration of training and therefore incurs significant communication costs as in previous VFL methods. 

To reduce the communication cost, FedBCD~\cite{liu2019communication} was proposed to leverage stale gradients for conducting local training such that participants can decrease the communication frequency from one local update to multiple updates. However, frequent iterative communication is still required for the whole training process. 
The other challenge is that the common samples across clients are usually limited in the real world, and training under such constraints may not achieve acceptable accuracy. FedCVT~\cite{kang2022fedcvt} proposes to expand the training samples by estimating representations and labels but does not address the bottleneck of communication cost.

In contrast, our proposed one-shot VFL and few-shot VFL are capable of solving the challenges of communication cost and limited common samples simultaneously.

\paragraph{Privacy in VFL.}

Privacy has become a concern of VFL since it was proposed because the clients need to send representations to the server for training, and privacy protection in VFL is well-explored. \cite{hardy2017private} presents a secure protocol that is managed by a third party, the coordinator, by employing privacy-preserving entity resolution and an additive homomorphic encryption scheme. To improve data privacy and model security, FATE~\cite{liu2021fate} applies a hybrid encryption scheme in the forward and backward stages of training. To defend the label inference attack, \cite{liu2021defending} proposes manipulating the labels following certain rules, which can be seen as a variant of label differential privacy (label DP)~\cite{chaudhuri2011sample, ghazi2021deep} in VFL. Our paper focuses on the performance and communication efficiency of VFL. However, our method does not require the clients and the server to share additional information compared with existing VFL methods and is orthogonal to existing privacy-protecting techniques, which can be directly applied to our method.

\paragraph{Semi-supervised Learning (SSL).}

SSL aims at training a model with partially labeled data, especially when the amount of labeled data is much smaller than the unlabeled ones. There have been many SSL algorithms proposed over the years. SSL algorithms can be broadly categorized as consistency regularization~\cite{bachman2014learning,rasmus2015semi}, pseudo-label methods~\cite{lee2013pseudo,miyato2018virtual,grandvalet2004semi,sohn2020fixmatch}, and generative models~\cite{kingma2013auto,doersch2016tutorial}. Consistency regularization is based on the assumption that if a realistic perturbation is applied to a data point, the prediction conducted by the trained model should not change significantly. MixMatch~\cite{berthelot2019mixmatch} applies consistency regularization along with entropy minimization and generic regularization and can achieve similar accuracy as fully supervised training approaches. Pseudo-labeling has become a component of many recent SSL techniques~\cite{miyato2018virtual}. Such methods leverage the trained model to generate pseudo labels for the unlabeled data so that the labeled training dataset is expanded. Generative models (e.g., VAE~\cite{kingma2013auto}) are trained to generate images from the data distribution and can be transferred to downstream tasks for a given task with targets.

Existing works~\cite{jeong2020federated,zhang2021improving,zhao2020semi,yang2021federated} apply SSL to FL to solve the real problem that the clients may not have enough labeled data. FSSL~\cite{jeong2020federated} learns inter-client consistency between multiple clients and splits model parameters for the server with labeled data and clients with unlabeled data separately. SemiFL~\cite{diao2021semifl} is the most recent work applying FixMatch~\cite{sohn2020fixmatch} to FL to improve the generalization of the global model. However, these methods focus on Horizontal Federated Learning (HFL), where the clients have the same feature space. Our work focuses on VFL settings where most clients have only partial features and no labels. In addition, existing deep SSL methods focus on imaging applications, while VFL has more potential for other types of data, such as tabular or multi-modal models combining imaging with other data types.

\section{Problem Definition}

Suppose $K$ clients and a server collaboratively train a model. There is an overlapped dataset\footnote{We assume the alignment between overlapping samples is known as a priori. In some applications private set intersection could be used before running VFL to find the sample alignment.} across all clients with size $N_o$:  $\{x_{o,i}, y_{o,i}\}_{i=1}^{N_o}$. The feature vector $x_{o,i}\in \mathbb{R}^{d}$ is distributed among $K$ clients $\{x_{o,i}^{k}\in \mathbb{R}^{d_k}\}_{k=1}^K$, where $d_k$ is the feature dimension of client $k$. For simplicity, the aligned dataset $\{x_{o,i}^{k}\in \mathbb{R}^{d_k}\}_{i=1}^{N_o}$ on client $k$ is denoted as $X_o^k$, and the set $\{X_o^k\}_{k\in[K]}$ is denoted as $X_o$. Besides $X_o^k$, each client $k$ also possesses $N_k$ local samples $\{x_{u,i}^{k}\in \mathbb{R}^{d_k}\}_{i=1}^{N_k}$ which is denoted as $X_u^k$ that is ``unaligned'' with other clients. The server has the true label of the overlapping samples $\{y_{o,i}\}_{i=1}^{N_o}$ which is denoted as $Y_o$. An example of data splitting in the two-client setting is shown in \cref{fig:data_split}. 

Each client (says the $k$-th) learns a representation extractor $f_k(.;\theta_k)$ to extract representations and the server learns a classifier $f_c(.;\theta_c)$ to classify the representations uploaded by clients. The collaborative training problem can be formulated as
%
%
{\small
\begin{equation}
    \min\limits_{\Theta} \mathcal{L}(\Theta;X_o, Y_o) \triangleq \frac{1}{N_o} \sum_{i=1}^{N_o} g(\theta_1,...,\theta_K,\theta_c;x_{o,i},y_{o,i}),
    \label{eq:problem_def}
\end{equation}
}
where $\Theta = \left[ \theta_1;...;\theta_K;\theta_c \right]$. $g(.)$ denotes the loss function formulated as:
\begin{equation}
    g(\theta_1,...,\theta_K,\theta_c;x_{o,i},y_{o,i}) = g\left(f_c\left(h_{o,i}^1\circ ... \circ h_{o,i}^K\right),y_{o,i}\right),
\end{equation}
where $\circ$ stands for the concatenation operation, and $h_{o,i}^k$ is the representation extracted by the local model on client $k$:
\begin{equation}
    h_{o,i}^k = f_k(x_{o,i}^k;\theta_k).
\end{equation}
For simplicity, the set of representations of the aligned data extracted by client $k$ $\{h_{o,i}^k\}_{i=1}^{N_o}$ is denoted as $H_o^k$. The objective of each party $k$ is to find the optimal $\theta_k$ without sharing local data $\{x_{o,i}^k\}_{i=1}^{N_o}$ and parameter $\theta_k$. The objective of the server is to optimize $\theta_c$ without sharing $\theta_c$ and true labels $Y_o$.

\section{Methods}

To reduce the communication cost and improve the model performance under the settings with limited overlapping users, 
we propose two VFL methods called one-shot VFL and few-shot VFL, respectively.

\subsection{One-shot Vertical Federated Learning}

\begin{figure}[ht]
\centering
     \includegraphics[scale=0.5]{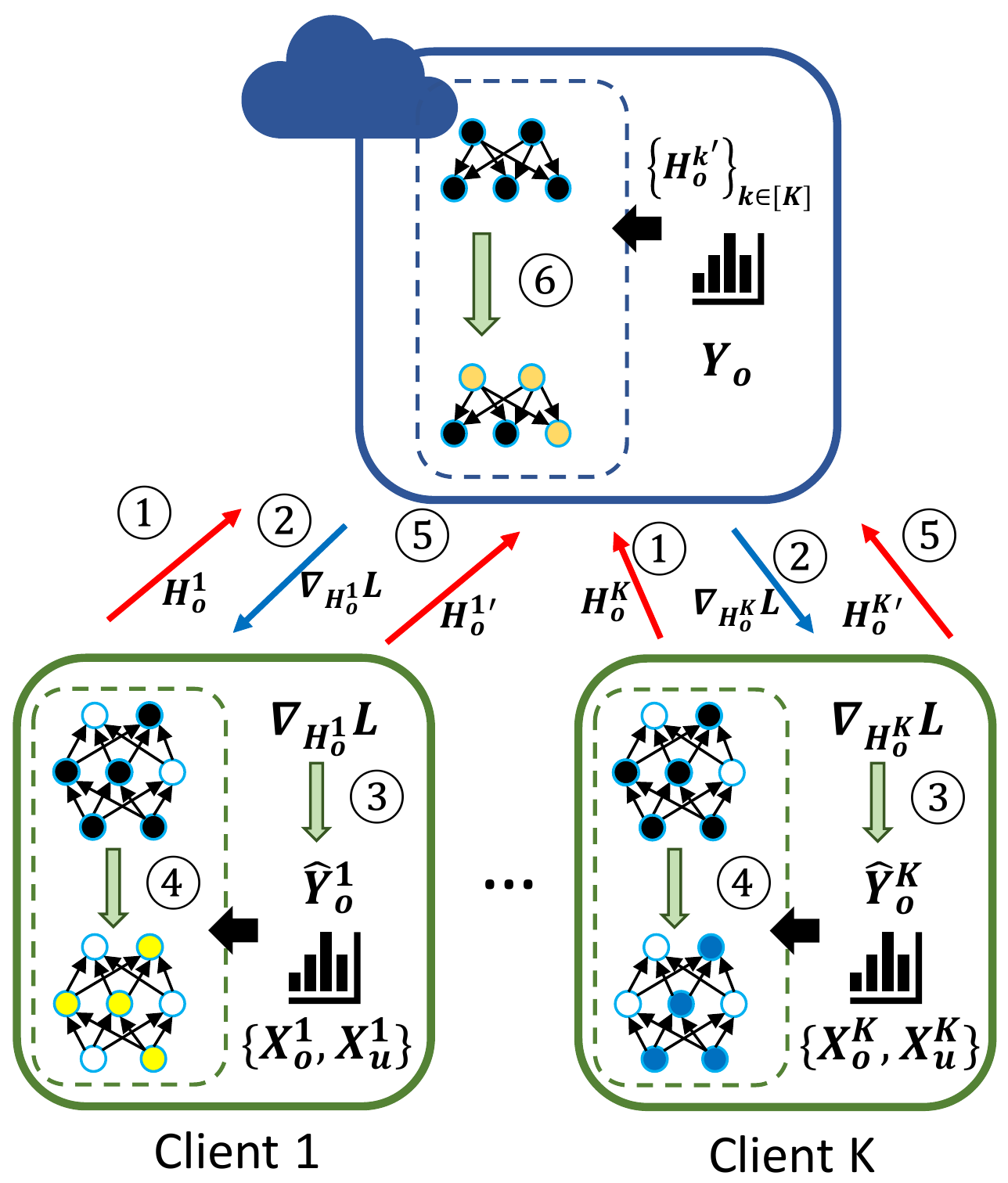}
\caption{Workflow of one-shot VFL. The clients conduct two times of uploading and one time of downloading.}
\label{fig:one-shot}
\end{figure}

We first propose one-shot VFL, in which the clients expect to receive partial gradients from the server only once. The intuition of one-shot VFL is that we can extract sufficient information that will guide clients to conduct local training from the received partial gradients. The workflow of one-shot VFL is shown in \cref{fig:one-shot}. First, the clients (e.g. the $k$-th) extract representations of the overlapping data $H_o^k$ and send the representations to the server (\circled{1}). Then, the server aligns and aggregates the received representations from all clients and computes loss with the true labels. After that, the server conducts back-propagation to compute the partial gradients of local representations $\nabla_{H_o^k} Loss$ and sends the partial gradients and the number of classes $C$ of the global classification task to corresponding (i.e., the $k$-th) clients (\circled{2}). After the $k$-th client receives the partial gradients, it conducts k-means on the partial gradients and assigns the overlapping samples $X^k_o$ temporary labels $\hat{Y}_o^k$ using the clustering index of corresponding gradients (\circled{3}). The intuition behind clustering is that the partial gradients of the same class should have similar directions while the partial gradients of different classes should have higher diversity. By clustering the partial gradients, the clients can infer information of true labels on the server to guide local training. With the temporary labels assigned to the overlapping samples, the clients conduct semi-supervised learning based on $X^k_u$ and $\{X^k_o,\hat{Y}_o^k\}$ to get updated $W_k'$ (\circled{4}). After the $k$-th client completes the local semi-supervised learning, it derives new representations ${H_o^k}'$ by computing $f_k(X^k_o;W_k')$ and sends ${H_o^k}'$ to the server (\circled{5}). Finally, the server aligns and aggregates the received new representations $\{{H_o^k}'\}_k$ to get new global representations $H_o'$ and finetunes the classifier $W_c$ using $\{H_o',Y_o\}$ (\circled{6}).

The detailed algorithm of one-shot VFL is shown in \cref{alg:one-shot}. It is notable that during the whole training process, the clients only need to upload representations to the server twice and download gradients from the server once, which is the reason we call it one-shot VFL. With such significant reduction from iterative download/upload to one-shot, we overcome the communication bottleneck in VFL. Meanwhile, local SSL conducted by clients fully utilizes the data of users that are unique to each client, and improves the performance under the realistic settings with limited overlapping samples.

\begin{algorithm}[ht] 
\footnotesize
\renewcommand{\algorithmicrequire}{\textbf{Server executes:}}
\renewcommand{\algorithmicensure}{\textbf{ClientUpdate($\nabla_{H_o^k}Loss, C$):}}
    \caption{\textbf{One-shot and few-shot VFL.} $mode$ is "few\_shot" if the server is executing few-shot VFL; $X_o^k$ and $X_u^k$ are aligned dataset and unaligned dataset of client $k$; $Y_o$ is the set of true labels of overlapping samples on the server; $C$ is the number of classes in the task; The $K$ clients are indexed by $k$; $B$ is the minibatch size; $E_s$ and $E_c$ are the number of epochs of the server and clients; $\eta_s$ and $\eta_c$ are learning rate of the server and clients; $g(.)$ and $l_{ssl}(.)$ are loss functions defined in \cref{eq:problem_def} and \cref{eq:loss_ssl}; Uploading happens in \textcolor{red}{$\gets$}; Downloading happens in \textcolor{blue}{$\gets$}.}
    \begin{algorithmic}[1] 
        \Require
        \State initialize $\theta_c$;
        \For{each client $k\in[K]$ \textbf{in parallel}}
            \State $H_o^k \textcolor{red}{\gets} f_k(X_o^k;\theta_k)$; \Comment{\circled{1} in \cref{fig:one-shot}}
        \EndFor
        \For{$k=1,...,K$}
            \State $\nabla_{H_o^k}Loss \textcolor{blue}{\gets} \nabla_{H_o^k} g\left(f_c\left(H_{o,i}^1\circ ... H_{o,i}^K\right),Y_{o}\right)$; \Comment{\circled{2} in \cref{fig:one-shot}}
        \EndFor
        \For{each client $k\in[K]$ \textbf{in parallel}}
            \State ClientUpdate($\nabla_{H_o^k}Loss, C$);\Comment{\circled{3}\circled{4} in \cref{fig:one-shot}}
            \State ${H_o^k} \textcolor{red}{\gets} f_k(X_o^k;\theta_k)$;\Comment{\circled{5} in \cref{fig:one-shot}}
            \If{$mode == $ \textit{"few\_shot"}}
                \State ${H_u^k} \textcolor{red}{\gets} f_k(X_u^k;\theta_k)$;
            \EndIf
        \EndFor
        \If{$mode == $ \textit{"few\_shot"}}
            \For{$k=1,...,K$}
                \State $\{\hat{p}_{u,i}^k\}_{i\in |H_u^k|} \textcolor{blue}{\gets} \text{InferProb}\left(\{{H_o^k}\}_k,{H_u^k}\right)$;\Comment{Defined in Alg.~\ref{alg:few-shot}}
                \State ClientUpdateFewshot($\{\hat{p}_{u,i}^k\}_{i\in |H_u^k|}$);\Comment{Defined in Alg.~\ref{alg:few-shot}}
                \State $H_o^k \textcolor{red}{\gets} f_k(X_o^k;\theta_k)$;
            \EndFor
        \EndIf
        \State $\mathcal{B_h, B_y} \gets$ (split $H_o^1\circ...\circ H_o^K$ and $Y_o$ into batches of size $B$);
        \For{epoch $i$ from 1 to $E_s$}
            \For{batch $b_h\in \mathcal{B_h}, b_y\in \mathcal{B_y}$}
                \State $\theta_c \gets \theta_c - \eta_s\nabla_{\theta_c} g\left(f_c\left(b_h\right),b_y\right)$;\Comment{\circled{6} in \cref{fig:one-shot}}
            \EndFor
        \EndFor
        
        \Ensure
        \State $\hat{Y}_o^k\gets \textit{k-means}(\nabla_{H_o^k}Loss, C)$; \Comment{\circled{3} in \cref{fig:one-shot}}
        \State $\mathcal{B_u}, \mathcal{B_o}, \mathcal{B_y} \gets$ (split $X^k_u, X^k_o, \hat{Y}_o^k$ into batches of size $B$);
        \For{epoch $i$ from 1 to $E_c$}
            \For{batch $b_u\in \mathcal{B_u}, b_o\in \mathcal{B_o}, b_y\in \mathcal{B_y}$}
                \State $\theta_k \gets \theta_k - \eta_c\nabla_{\theta_k} l_{ssl}\left(\theta_k;b_u, b_o, b_y\right)$; \Comment{\circled{4} in \cref{fig:one-shot}}
            \EndFor
        \EndFor
    \end{algorithmic}
    \label{alg:one-shot}
\end{algorithm}

%
\paragraph{Local SSL.}
In one-shot VFL, the clients conduct semi-supervised learning (SSL) based on $X^k_u$ and $\{X^k_o,\hat{Y}_o^k\}$. For different types of data, the detailed SSL algorithms are different. The training objective of the $k$-th client can be abstracted into 
\begin{equation}
    l_{ssl}\left(\theta_k;X^k_u, X^k_o, \hat{Y}_o^k\right) = l_s\left(\theta_k;X^k_o, \hat{Y}_o^k\right) + \lambda_ul_{u}\left(\theta_k;X^k_u\right),
    \label{eq:loss_ssl}
\end{equation}
where $l_s(.)$ is the supervised training loss, $l_{u}(.)$ is the unsupervised training loss, $\lambda_u$ controls the trade-off between the supervised loss and unsupervised loss. In this paper, we focus on two types of data: image data and tabular data, which are common use cases of VFL. Plenty of SSL algorithms~\cite{sohn2020fixmatch,berthelot2019mixmatch} have been proposed for image recognition, and we apply FixMatch~\cite{sohn2020fixmatch} for the clients conducting SSL on image data, which is a widely applied SSL algorithm in image recognition. On the other hand, in order to fit DL to the task of tabular data SSL, we modify the augmentation methods in FixMatch and propose our FixMatch-tab algorithm for local SSL of tabular data. 
\begin{figure}[ht]
\centering
     \includegraphics[scale=0.3]{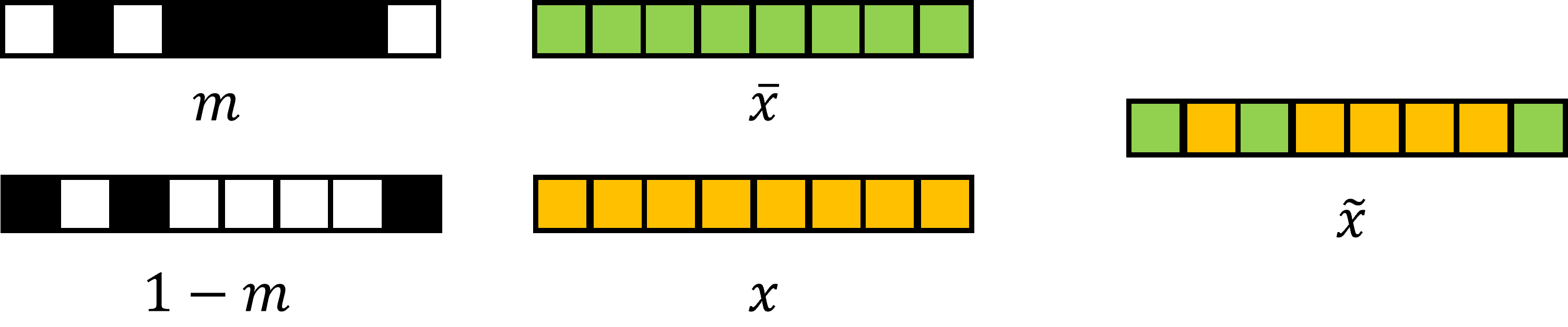}
\caption{Features are randomly masked for data augmentation.}
\label{fig:mask}
\end{figure}
We modify the weak augmentation $\alpha(.)$ and strong augmentation $\mathcal{A}(.)$ in FixMatch to adapt it to the tabular data. For weak augmentation, we randomly generate a binary mask $m$ with the same shape of the data point. Each element of $m$ is sampled from a Bernoulli distribution. We replace the masked elements with the mean value of the corresponding elements of local data. For the strong augmentation, we add noise to the masked samples. Thus, when we train a data point $x$ in FixMatch-tab, we first sample a binary mask $m$ for both weak and strong augmentation and sample a noise vector $n$ for strong augmentation where
{\small
\begin{equation}
\begin{aligned}
    &m_i \sim B(1,r_m),\\
    &n_i \sim N(0,\sigma^2),
\end{aligned}
\end{equation}
}%
$r_m$ is the expected ratio of elements that are masked and $\sigma^2$ is variance of the noise. Then, the weak augmentation $\alpha$ and strong augmentation $\mathcal{A}(.)$ for this sample in FixMatch-tab are formulated as 
{\small
\begin{equation}
\begin{aligned}
    &\alpha\left(x\right) = m\otimes x + (1-m)\otimes \bar{x},\\
    &\mathbb{A}\left(x\right) = \alpha\left(x\right) + n,
\end{aligned}
\end{equation}
}%
where $\bar{x}=\frac{1}{N}\sum_{j\in [N]}x_j$ and $N$ is the number of local samples.
%
%
%
%
%
\subsection{Few-shot Vertical Federated Learning}
\begin{figure}[ht]
\centering
     \includegraphics[scale=0.4]{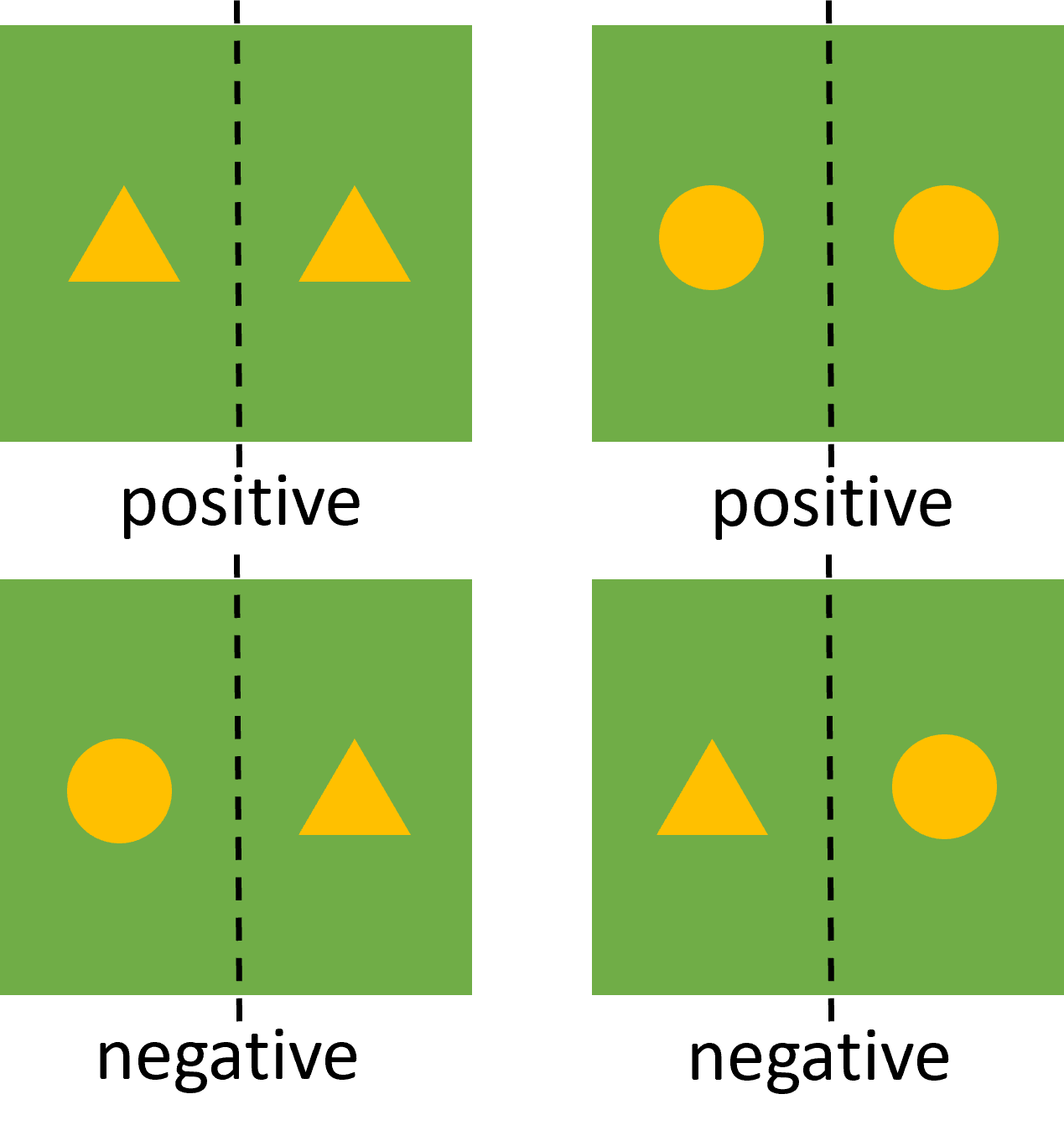}
\caption{A client does not have enough information to generate reasonable pseudo labels from either the left or right part of the image alone during local semi-supervised learning (SSL).}
\label{fig:toy_example}
\end{figure}

\begin{figure*}[ht]
\centering
     \includegraphics[scale=0.4]{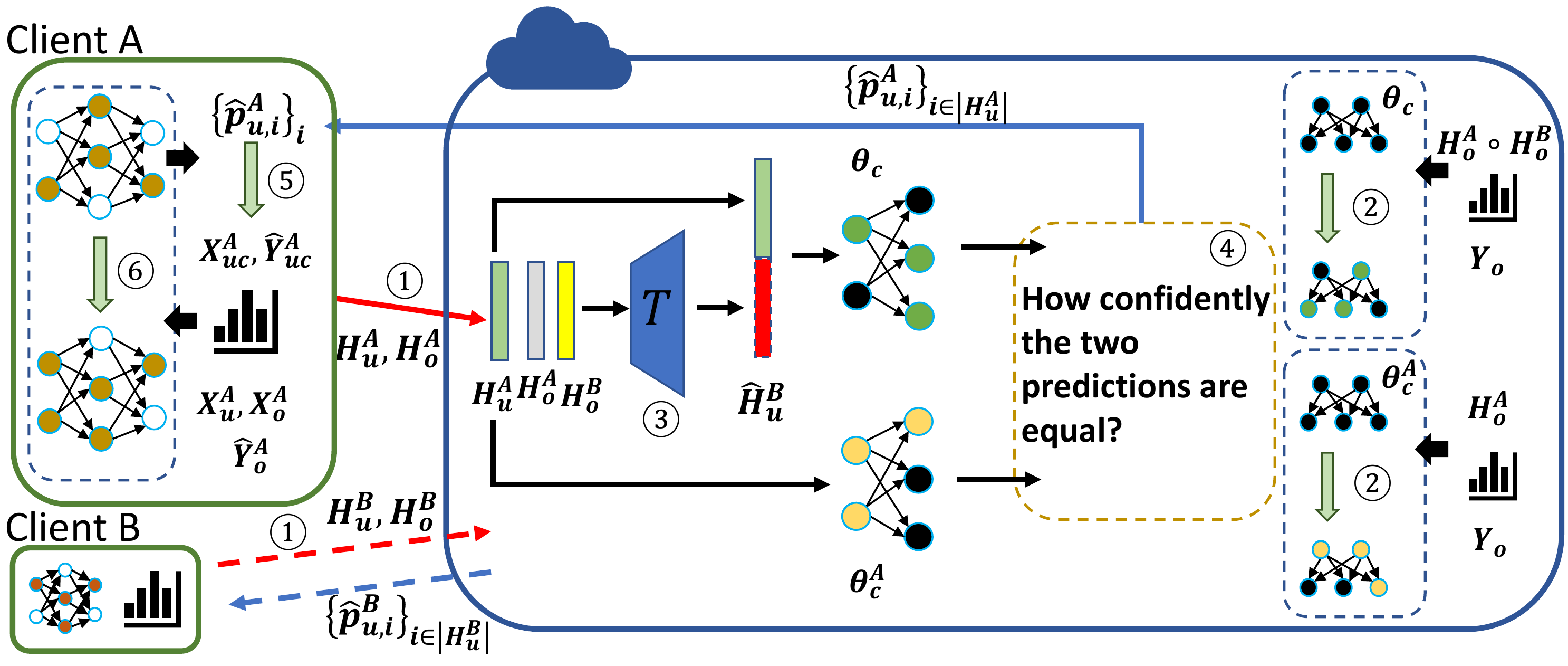}
\caption{The server judges whether local samples contain enough information to generate accurate pseudo labels in few-shot learning. The clients conduct local SSL with the expanded labeled dataset.}
\label{fig:few_shot_server}
\end{figure*}

\begin{algorithm}[ht] 
\footnotesize
\renewcommand{\algorithmicrequire}{\textbf{ClientUpdateFewshot($\{\hat{p}_{u,i}^k\}_{i\in |X_u^k|}$):}}
\renewcommand{\algorithmicensure}{\textbf{InferProb$\left(\{{H_o^k}\}_k,{H_u^k}\right)$:}}
    \caption{Local SSL training process with expanded labeled dataset in few-shot VFL. $H_o^{[K]\backslash\{k\}}$ stands for $H_o^1\circ,...,\circ H_o^{k-1}\circ H_o^{k+1}\circ,...,\circ H_o^K$ }
    \begin{algorithmic}[1] 
        \Ensure
        \For{$k=1,...,K$} \Comment{This is executed only by once}
            \State $\theta_c^k \gets \argmin_{\theta} g\left(f_c^k\left(H_o^k;\theta\right), Y_o\right)$; \Comment{Optimize with SGD, \circled{2} in \cref{fig:few_shot_server}}
        \EndFor
        \State $\theta_c \gets $ execute line 22-27 in Alg.~\ref{alg:one-shot}; \Comment{\circled{2} in \cref{fig:few_shot_server}}
        \State $\hat{H}_u^{[K]\backslash\{k\}} \gets T \left(H_u^k, H_o^k, H_o^{[K]\backslash\{k\}}\right)$; \Comment{\circled{3} in \cref{fig:few_shot_server}}
        \For{$i = 1,...,|H_u^k|$}
            \State $\hat{p}_{u,i}^k \gets$ compute \cref{eq:probs} and \cref{eq:pseudo_probs}; \Comment{\circled{4} in \cref{fig:few_shot_server}}
        \EndFor
        \State \Return $\{\hat{p}_{u,i}^k\}_{i\in |H_u^k|}$;
        
        \Require
        \State $X_{uc}^k \gets \text{sample from } X_u^k \text{ with probability } \{\hat{p}_{u,i}^k\}_{i\in |X_u^k|}$; \Comment{\circled{5} in \cref{fig:few_shot_server}}
        \State $\hat{Y}_{uc}^k \gets f_k\left( X_{uc}^k;\theta_k\right)$; \Comment{\circled{5} in \cref{fig:few_shot_server}};
        \State $X_o^{k\prime}\gets X_o^k \cup X_{uc}^k$; ${\hat{Y}_o}^{k\prime}\gets \hat{Y}_o^k \cup \hat{Y}_{uc}^k$;
        \State ${X_u^k}'\gets X_o^k \backslash X_{uc}^k$;
        \State $\mathcal{B_u}, \mathcal{B_o}, \mathcal{B_y} \gets$ (split $X_u^{k\prime}, X_o^{k\prime}, \hat{Y}_o^{k\prime}$ into batches of size $B$);
        \For{epoch $i$ from 1 to $E_c$}
            \For{batch $b_u\in \mathcal{B_u}, b_o\in \mathcal{B_o}, b_y\in \mathcal{B_y}$}
                \State $\theta_k \gets \theta_k - \eta_c\nabla_{\theta_k} l_{ssl}\left(\theta_k;b_u, b_o, b_y\right)$; \Comment{\circled{6} in \cref{fig:few_shot_server}}
            \EndFor
        \EndFor
        \end{algorithmic}
    \label{alg:few-shot}
\end{algorithm}

Even though one-shot VFL can achieve high performance of the global model with extremely low communication cost, we consider improving the performance further by paying with a few more rounds of communication. The key factor to improve SSL performance is to have a larger labeled dataset. Thus, we propose to expand the supervised learning dataset on clients in VFL. One intuitive idea is to assign pseudo labels from local predictions to the unlabeled data points if those predictions have a high confidence. By doing this, we would only need to modify the local training procedure of one-shot VFL without introducing additional communication cost. However, there is one potential problem this method cannot solve. Considering a toy example shown in \cref{fig:toy_example}. Two clients participate in training a image classification task, and each client has access to a half of each image. If the two shapes on the image are the same, the image is positive, and negative if both shapes are different. If we want to improve the performance by enlarging the labeled dataset, the key is to generate pseudo labels with high accuracy. However, in this toy example, it is impossible for the clients to generate reasonable pseudo labels based on only half of the images since they do not have enough information to infer the true label of the image.

To solve this problem, we propose few-shot VFL as shown in \cref{alg:one-shot}. The main difference with one-shot VFL is that the server in few-shot VFL estimates the missing part of the representations for each client's unaligned data which shall expand their labeled datasets. The detailed pipeline of line 17-18 in \cref{alg:one-shot} is shown in \cref{fig:few_shot_server}. In this section we focus on VFL with two clients (A \& B), but the proposed method can be naturally extended to the scenario where there are more than two clients. When the server receives the representations $H_{u}^A$ of client A's unaligned data (\circled{1} in \cref{fig:few_shot_server}), it estimates the missing representations  $\hat{H}_u^B$ of corresponding samples on client B (does not exist for unaligned data) with a transform layer $T$ (\circled{3} in \cref{fig:few_shot_server}) as
{\small
\begin{equation}
    \hat{H}_u^B = T(H_{u}^A, H_{o}^A, H_{o}^B).
\end{equation}
}%
The details of $T(.)$ will be introduced later. Then for each unaligned sample (says the $i$-th), the server produces predictions $\{\hat{y}_{u,i}^A, \hat{y}_{u,i}^{A,B}\}$ and probabilities $\{p_{u,i}^A, p_{u,i}^{A,B}\}$ following 
{\small
\begin{equation}
    \begin{aligned}
        \hat{y}_{u,i}^A &= \argmax_{j} f_c^A(H_{u,i}^A;\theta_c^{A})_j,\\
        p_{u,i}^A &= \max_{j} f_c^A(H_{u,i}^A;\theta_c^{A})_j,\\
        \hat{y}_{u,i}^{A,B} &= \argmax_{j} f_c(H_{u,i}^{A}\circ \hat{H}_{u,i}^{B};\theta_c^{A,B})_j,\\
        p_{u,i}^{A,B} &= \max_{j} f_c(H_{u,i}^{A}\circ \hat{H}_{u,i}^{B};\theta_c^{A,B})_j.
    \label{eq:probs}
    \end{aligned}
\end{equation}
}%
where $f_c^A(.)$ is an auxiliary classifier whose input is $h_{u,i}^A$. $\theta_c^{A}$ and $\theta_c^{A,B}$ are trained (\circled{2} in \cref{fig:few_shot_server}) based on the overlapping samples. If the predictions $\{\hat{y}_{u,i}^A, \hat{y}_{u,i}^{A,B}\}$ based on local and estimated global representations are the same with high confidence, the local representation $h_{u,i}^A$ contains enough information and should be given a pseudo label during the local SSL on client A. To reduce the noise from misleading pseudo labels, the server sets a probability $\hat{p}_{u,i}^A$ for each unaligned sample to be given a pseudo-label during local training following
{\small
\begin{equation}
    \hat{p}_{u,i}^A = \mathbb{1}\left(\hat{y}_{u,i}^A = \hat{y}_{u,i}^{A,B}\right)\mathbb{1}\left(p_{u,i}^A > t\right)\mathbb{1}\left(p_{u,i}^{A,B} > t\right)p_{u,i}^{A,B}.
    \label{eq:pseudo_probs}
\end{equation}
}%
The intuition of $\hat{p}_{u,i}^A$ is that with the higher confidence the local representation $h_{u,i}^A$ is predicted the same label with the global representation $h_{u,i}^A\circ\hat{h}_{u,i}^B$, the larger probability $\hat{p}_{u,i}^A$ with which the $i$-th unaligned sample on client A should be given a pseudo label during local training. In the following, we will introduce the representation transform layer $T(.)$ and local SSL with probability set $\{\hat{p}_{u,i}^A\}_i$.
\paragraph{Efficient Representation Estimation.} We design the representation transform layer utilizing a \textit{scaled dot product attention} (SDPA) function formulated as
{\small
\begin{equation}
\begin{aligned}
    \hat{H}_u^B &= T(H_{u}^A, H_{o}^A, H_{o}^B)\\
    &= softmax(\frac{H_{u}^A \otimes {H_{o}^A}^T}{\sqrt{d}})\otimes{H_{o}^B},
\end{aligned}
\end{equation}
}%
where $\otimes$ is matrix multiplication operator, and $d$ is the dimension of representation. With $T(.)$, Each missing representation is estimated through the weighted sum over representations of overlapped samples. The weight matrix $W_A=softmax(\frac{H_{u}^A \otimes {H_{o}^A}^T}{\sqrt{d}})$ reflects the similarity between the representation to be estimated and the aligned representations in client A.

We apply the SDPA function rather than a generative model (e.g., GAN) to estimate representations for two reasons. First, the generative model has to be trained on the representations of overlapping samples. However, the amount of overlapping samples could be too small in real life to train a generator with good performance. Second, when there are $K$ clients, the server needs to train $K$ generators, which introduces heavy computational overhead. By applying the SDPA function to estimate representations, we can overcome the problem of limited overlapping samples and improve the computational efficiency of our estimation.
\paragraph{Local SSL with Expanded Supervised Dataset.}
After the client A receives $\{\hat{p}_{u,i}^A\}$ (\circled{4} in \cref{fig:few_shot_server}), it samples a subset denoted as $X_{uc}^A$ from $X_u^A$ with probabilities $\{\hat{p}_{u,i}^A\}$. For each sample $x_{uc,i}^A$ in $X_{uc}^A$, client A assigns the pseudo label $\hat{y}_{uc,i}^A$ as the prediction of the local model that was learned using SSL on client A. For simplicity, the set of pseudo labels $\hat{y}_{uc,i}^A$ is denoted as $\hat{Y}_{uc}^A$. In such way, client A expands the supervised data via SSL. Hence, the objective of SSL (\circled{6} in \cref{fig:few_shot_server}) on client A can be formulated as
{\small
\begin{equation}
    \begin{aligned}
    &l_{ssl}\left(\theta;X^A_u\backslash X_{uc}^A, X^A_o\cup X_{uc}^A, \hat{Y}_o^A\cup \hat{Y}_{uc}^A\right) \\
    = &l_s\left(\theta;X^A_o\cup X_{uc}^A, \hat{Y}_o^A\cup \hat{Y}_{uc}^A\right) + \lambda_ul_{u}\left(\theta;X^A_u\backslash X_{uc}^A\right).
    \end{aligned}
\end{equation}
}

\section{Evaluation}

\subsection{Experimental setup}
We evaluate our proposed one-shot VFL and few-shot VFL on both image and tabular data. As stated before, we focus and evaluate on two-client scenarios in this paper, which is a common experimental setup in most VFL literature\cite{liu2019communication,liu2020federated}. We compare our methods with two SOTA VFL methods aiming at reducing communication cost and solving the problem of limited overlapping samples, respectively.
\vspace{-3mm}
\paragraph{Baselines.} We compare our proposed algorithm with vanilla VFL and two SOTA VFL methods. (1) \textbf{FedBCD~\cite{liu2019communication}} aims at reducing the communication cost. In vanilla VFL, clients conduct one iteration of training after one time of inter-party communication. FedBCD allows clients to conduct multiple iterations of local training using the stale partial gradients of representations received in the last communication. (2) \textbf{FedCVT~\cite{kang2022fedcvt}} is a semi-supervised learning approach that improves the performance of VFL using limited overlapping samples. FedCVT leverages representation estimation and pseudo-labels prediction to expand the training set to improve the model’s representation learning. However, it still suffers from high communication cost.


%
\vspace{-3mm}
\paragraph{Datasets.} To evaluate our VFL methods under different VFL settings, we use both image data and tabular data for experiments. We use CIFAR-10 for image classification and UCI\_credit\_card dataset~\cite{yeh2009comparisons} for prediction of default of credit card clients. For CIFAR-10, we follow~\cite{liu2019communication,kang2022fedcvt} to split an image into two halves. For UCI\_credit\_card dataset, we follow FATE~\cite{liu2021fate} to assign ten attributes to one client and the rest to the other client. To mimic the settings that limited samples are overlapping, we randomly sample $N_o$ samples from the dataset as the aligned dataset. For the rest of samples we evenly and randomly separate them into two sets, and one client has access to the assigned attributes/halves of images of one set.
\vspace{-3mm}
\paragraph{Hyperparameter Configurations.}
To evaluate our methods under settings with different sizes of overlapping samples, we set $N_o=\{256, 512, 1024, 2048\}$ for CIFAR-10 and $N_o=\{1000, 2000\}$ for UCI\_default\_credit. We set $B$ as 32 for both datasets. Learning rates $\eta_s$ and $\eta_c$ are set to be 0.01. For FedBCD, we set $Q$ as 5. We set $\sigma$ as 0.1 and $r_m$ as 0.2 for tabular data augmentation. For CIFAR-10, we allow the baselines to continue training even after convergence to try to achieve a decent accuracy. For UCI\_default\_credit, we stop the training of baselines where there is no improvement on accuracy in the last 20 rounds. We use WideResNet20 as the backbone model for CIFAR-10 and a two-layer MLP for UCI\_default\_credit.

\vspace{-3mm}
\paragraph{Evaluation metrics.} (1) \textbf{Utility metric (Accuracy \& AUC):} We use the test data accuracy of the classifier on the server to measure the performance of VFL on CIFAR-10. For UCI\_default\_credit, we apply Area under the ROC Curve (AUC) as the utility metric. A smaller accuracy or AUC means a less practical utility. (2) \textbf{Communication metric (Communication cost/times):} We use the times of communication between the clients and the server and the total data volume of communication cost to evaluate the communication efficiency of VFL.
\subsection{Experimental results}
\begin{table*}[th]
\small
    \centering
    \caption{Results of accuracy and communication on CIFAR-10. Best accuracy is shown in \textbf{bold} and best communication cost in \textbf{\textit{bold italic}}.}
        \begin{tabular}{l || c  c c| c  c c| c  c c| c  c c}
            \toprule
            \textbf{Overlap size} & \multicolumn{3}{c|}{256} & \multicolumn{3}{c|}{512} &\multicolumn{3}{c|}{1024} &\multicolumn{3}{c}{2048}\\
            \hline
            & \tabincell{c}{Acc\\(\%)} & \tabincell{c}{Comm \\times} & \tabincell{c}{Comm \\cost\\(MB)} & \tabincell{c}{Acc\\(\%)} & \tabincell{c}{Comm \\times} & \tabincell{c}{Comm \\cost\\(MB)} & \tabincell{c}{Acc\\(\%)} & \tabincell{c}{Comm \\times} & \tabincell{c}{Comm \\cost\\(MB)} & \tabincell{c}{Acc\\(\%)} & \tabincell{c}{Comm \\times} & \tabincell{c}{Comm \\cost\\(MB)} \\
            \hline
            Vanilla VFL & 31.47 & 8000 & 262 & 35.33 & 16000 & 524 & 42.71 & 32000 & 1047 & 50.75 & 64000 & 2094\\
            \hline
            FedCVT~\cite{kang2022fedcvt} & 31.83 & 8000 & 262 & 35.12 & 16000 & 524 & 42.38 & 32000 & 1047 & 48.2 & 64000 & 2094\\
            \hline
            FedBCD~\cite{liu2019communication} & 31.45 & 1600 & 53 & 35.43 & 3200 & 105 & 41.93 & 6400 & 209 & 49.75 & 12800 & 419 \\
            \hline
            One-shot VFL & 78.23 & \textbf{\textit{3}} & \textbf{\textit{0.79}} & 81.12 & \textbf{\textit{3}} & \textbf{\textit{1.6}} & 85.25 & \textbf{\textit{3}} & \textbf{\textit{3.1}} & 86.13 & \textbf{\textit{3}} & \textbf{\textit{6.3}} \\
            \hline
            Few-shot VFL & 78.93 & 5 & 26.4 & 83.03 & 5 & 27.2 & 85.68 & 5 & 28.7 & 87.23 & 5 & 31.9 \\
            \hline
            \tabincell{l}{Few-shot VFL \\+finetune} & \textbf{80.37} & 805 & 52.6 & \textbf{84.05} & 965 & 58.6 & \textbf{86.35} & 1805 & 87.7 & \textbf{87.49} & 2005 & 97.4\\
            \bottomrule
        \end{tabular}
    \label{tb:results_cifar}
\end{table*}


\paragraph{Accuracy v.s. Communication Cost.}
The results of CIFAR-10 are shown in \cref{tb:results_cifar}. It is shown that compared with vanilla VFL, one-shot VFL improves the accuracy by more than 45\% while reducing communication cost by more than 330$\times$. FedBCD reduces communication cost compared with Vanilla VFL. However, it does not improve the accuracy of the model, and the communication reduction is not comparable with one-shot VFL. It is notable that FedCVT cannot achieve significant accuracy improvement compared with vanilla VFL, because the true label is extremely limited in our realistic setting. With extremely limited true labels, it is hard for the server of FedCVT to conduct SSL using only the estimated representations and pseudo labels. 

Few-shot VFL improves the accuracy further with higher communication cost compared with one-shot VFL. However, the communication reduction is still significant compared with baselines. In both one-shot VFL and few-shot VFL, clients train representation extractors first, then the server trains the classifier. To improve the accuracy further, we conduct end-to-end vanilla VFL after completing few-shot VFL to finetune the global model on CIFAR-10. It is shown that the end-to-end finetuning can improve the accuracy further. Even though the finetuning requires more communication rounds, it is still much more efficient compared with the baselines and offers the clients an option to further improve the performance.

\begin{figure}[ht]
\centering
     \includegraphics[scale=0.34]{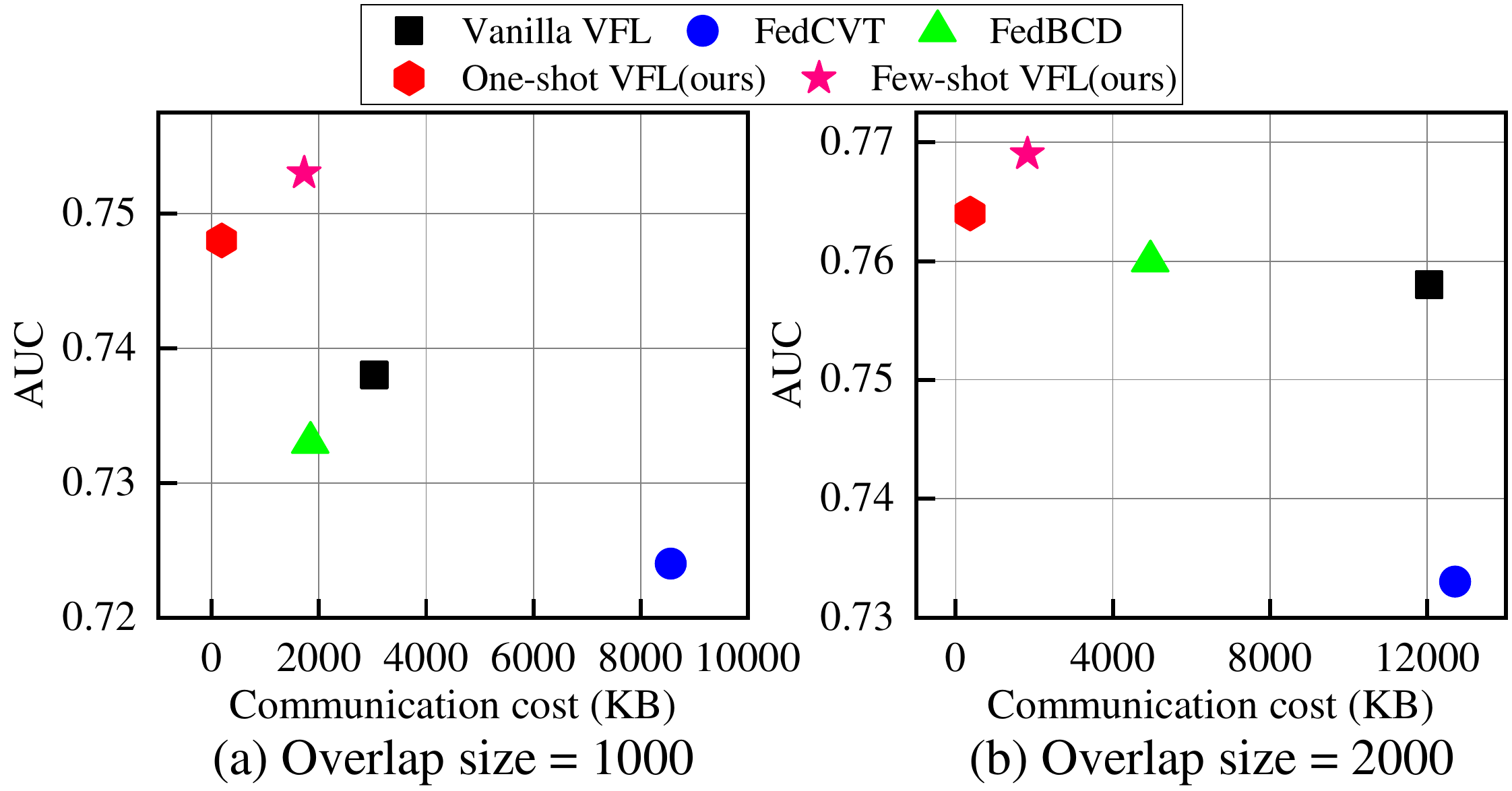}
\caption{Compared results of AUC v.s. communication cost on UCI\_default\_credit.}
\label{fig:tab-cost}
\end{figure}

The results of UCI\_default\_credit dataset are shown in \cref{fig:tab-cost}. Even though the task of credit card default detection is much simpler than image classification, which does not require a large amount of data to learn, our one-shot VFL can still achieve AUC higher than all the baselines in both settings. One-shot VFL can reduce the communication cost by more than 32$\times$, 33$\times$ and 10$\times$ compared with Vanilla VFL, FedCVT and FedBCD, respectively, under the setting with 2000 overlapping samples. Few-shot VFL increases the communication cost slightly compared with one-shot VFL, but it can improve the AUC further.

\begin{figure}[ht]
\centering
     \includegraphics[scale=0.34]{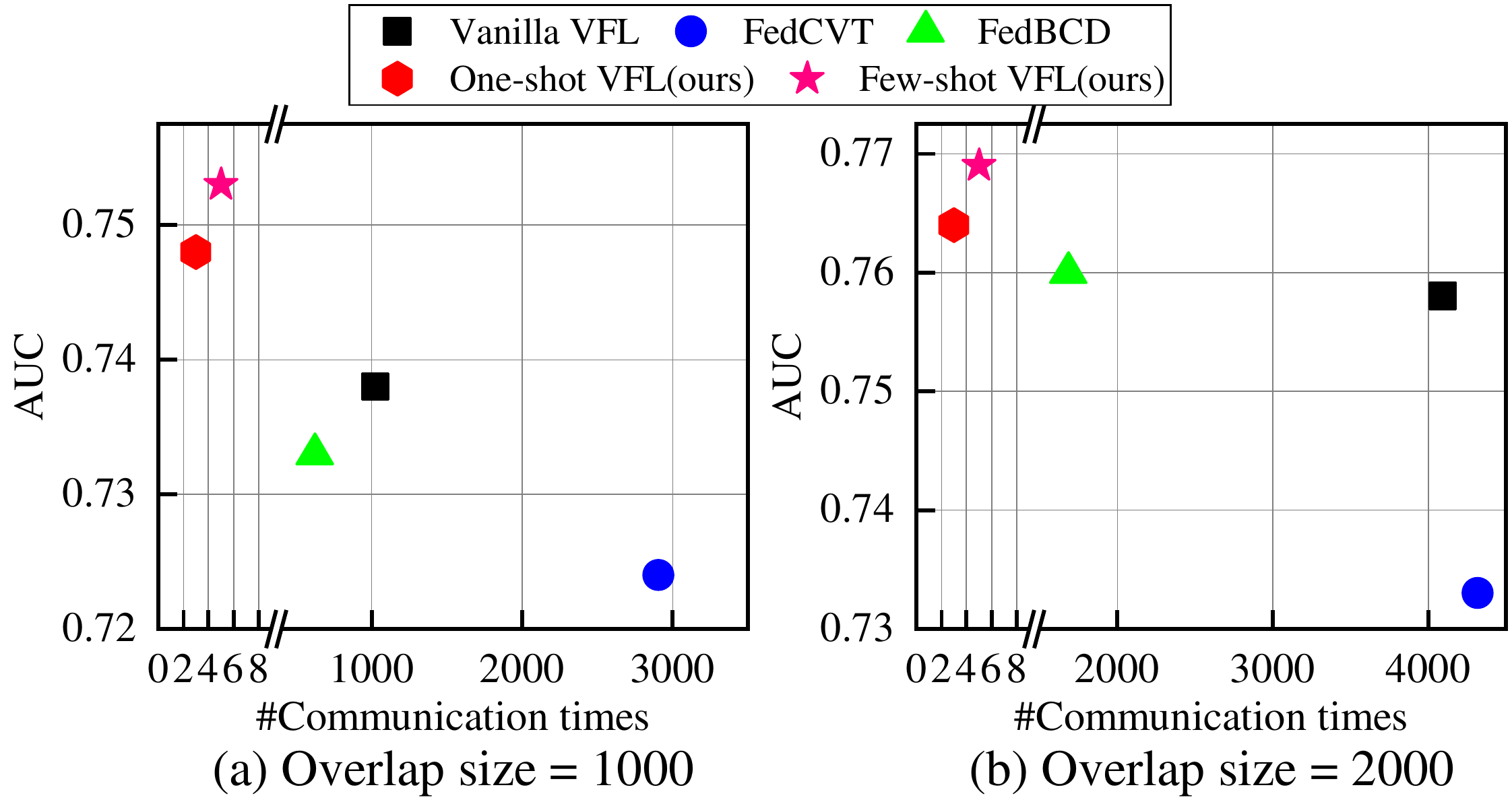}
\caption{Compared results of AUC v.s. number of communication times on UCI\_default\_credit.}
\label{fig:tab-time}
\end{figure}

%
\vspace{-3.5mm}
\paragraph{Accuracy v.s. Times of Communication.}
Besides the communication cost, the times of communication needed between the clients are also an important metric to evaluate the communication efficiency. If the clients are required to communicate with the server continually (e.g., vanilla VFL), a stable and reliable communication channel between the server and a client will be necessary. In addition, if a client cannot upload its update to the server in one round of training due to network outage, all the other clients have to wait for it, which is extremely detrimental to the robustness and efficiency of the system. As shown in \cref{tb:results_cifar} and \cref{fig:tab-time}, only three times of communications are needed for one-shot VFL, and the clients conduct SSL locally without waiting for the response from the server, which improves the efficiency significantly. FedBCD reduces the times of communication, but it is still not comparable to one-shot and few-shot VFL. In addition, continual communication between clients are still required for FedBCD, which cannot solve the bottleneck of communication efficiently. Few-shot VFL improves the accuracy further with only two additional times of communication between the clients and the server. During finetuning, the clients need to continually communicate with the server for multiple rounds. However, the clients do not need to communicate during as many rounds as for the other baselines. In practice, our one-shot and few-shot VFL can be used as pre-training techniques to achieve higher performance while significantly reducing the communication cost.

\section{Conclusions and Future Work}

In this paper, we propose one-shot VFL that applies SSL to solve two critical problems of VFL: high communication cost and limited overlapping samples common in the real world. In one-shot VFL, the clients only conduct two times of uploading and one time of downloading and can achieve higher accuracy than the SOTA VFL approaches. We also propose few-shot VFL to improve the performance further by paying with one more round of communication. We evaluate our methods on imaging data and tabular data, and the results demonstrate that our methods can improve the model performance under the settings with limited overlapping samples and reduce the communication cost significantly.

In the future, we will evaluate our VFL methods in multi-modal settings combining different data types. In addition, we note that data privacy preservation is a significant concern of deploying FL in real life~\cite{sun2022label,wu2020privacy}. Our methods do not require the clients and the server to share additional information compared with existing VFL methods besides the number of classes, which is common knowledge for both the server and clients in most settings. 
Existing defense methods~\cite{liu2021defending,ghazi2021deep} can be directly incorporated into our approaches to improve privacy. In this paper, we follow most previous VFL literature~\cite{liu2019communication,liu2021fate} and evaluate on two-client scenarios. We will explore multi-party settings in future work.
Our code will be made publicly available.

\clearpage
\newpage
{\small
\bibliographystyle{ieee_fullname}
\bibliography{reference}

\begin{thebibliography}{10}\itemsep=-1pt

\bibitem{bachman2014learning}
Philip Bachman, Ouais Alsharif, and Doina Precup.
\newblock Learning with pseudo-ensembles.
\newblock {\em Advances in neural information processing systems}, 27, 2014.

\bibitem{berthelot2019mixmatch}
David Berthelot, Nicholas Carlini, Ian Goodfellow, Nicolas Papernot, Avital
  Oliver, and Colin~A Raffel.
\newblock Mixmatch: A holistic approach to semi-supervised learning.
\newblock {\em Advances in neural information processing systems}, 32, 2019.

\bibitem{chaudhuri2011sample}
Kamalika Chaudhuri and Daniel Hsu.
\newblock Sample complexity bounds for differentially private learning.
\newblock In {\em Proceedings of the 24th Annual Conference on Learning
  Theory}, pages 155--186. JMLR Workshop and Conference Proceedings, 2011.

\bibitem{cheng2021secureboost}
Kewei Cheng, Tao Fan, Yilun Jin, Yang Liu, Tianjian Chen, Dimitrios
  Papadopoulos, and Qiang Yang.
\newblock Secureboost: A lossless federated learning framework.
\newblock {\em IEEE Intelligent Systems}, 36(6):87--98, 2021.

\bibitem{diao2021semifl}
Enmao Diao, Jie Ding, and Vahid Tarokh.
\newblock Semifl: Communication efficient semi-supervised federated learning
  with unlabeled clients.
\newblock {\em arXiv preprint arXiv:2106.01432}, 2021.

\bibitem{doersch2016tutorial}
Carl Doersch.
\newblock Tutorial on variational autoencoders.
\newblock {\em arXiv preprint arXiv:1606.05908}, 2016.

\bibitem{fu2021vf2boost}
Fangcheng Fu, Yingxia Shao, Lele Yu, Jiawei Jiang, Huanran Xue, Yangyu Tao, and
  Bin Cui.
\newblock Vf2boost: Very fast vertical federated gradient boosting for
  cross-enterprise learning.
\newblock In {\em Proceedings of the 2021 International Conference on
  Management of Data}, pages 563--576, 2021.

\bibitem{ghazi2021deep}
Badih Ghazi, Noah Golowich, Ravi Kumar, Pasin Manurangsi, and Chiyuan Zhang.
\newblock Deep learning with label differential privacy.
\newblock {\em Advances in neural information processing systems},
  34:27131--27145, 2021.

\bibitem{grandvalet2004semi}
Yves Grandvalet and Yoshua Bengio.
\newblock Semi-supervised learning by entropy minimization.
\newblock {\em Advances in neural information processing systems}, 17, 2004.

\bibitem{hard2018federated}
Andrew Hard, Kanishka Rao, Rajiv Mathews, Swaroop Ramaswamy, Fran{\c{c}}oise
  Beaufays, Sean Augenstein, Hubert Eichner, Chlo{\'e} Kiddon, and Daniel
  Ramage.
\newblock Federated learning for mobile keyboard prediction.
\newblock {\em arXiv preprint arXiv:1811.03604}, 2018.

\bibitem{hardy2017private}
Stephen Hardy, Wilko Henecka, Hamish Ivey-Law, Richard Nock, Giorgio Patrini,
  Guillaume Smith, and Brian Thorne.
\newblock Private federated learning on vertically partitioned data via entity
  resolution and additively homomorphic encryption.
\newblock {\em arXiv preprint arXiv:1711.10677}, 2017.

\bibitem{hu2019learning}
Yaochen Hu, Peng Liu, Linglong Kong, and Di Niu.
\newblock Learning privately over distributed features: An admm sharing
  approach.
\newblock {\em arXiv preprint arXiv:1907.07735}, 2019.

\bibitem{jeong2020federated}
Wonyong Jeong, Jaehong Yoon, Eunho Yang, and Sung~Ju Hwang.
\newblock Federated semi-supervised learning with inter-client consistency \&
  disjoint learning.
\newblock {\em arXiv preprint arXiv:2006.12097}, 2020.

\bibitem{kairouz2021advances}
Peter Kairouz, H~Brendan McMahan, Brendan Avent, Aur{\'e}lien Bellet, Mehdi
  Bennis, Arjun~Nitin Bhagoji, Kallista Bonawitz, Zachary Charles, Graham
  Cormode, Rachel Cummings, et~al.
\newblock Advances and open problems in federated learning.
\newblock {\em Foundations and Trends{\textregistered} in Machine Learning},
  14(1--2):1--210, 2021.

\bibitem{kang2022fedcvt}
Yan Kang, Yang Liu, and Xinle Liang.
\newblock Fedcvt: Semi-supervised vertical federated learning with cross-view
  training.
\newblock {\em ACM Transactions on Intelligent Systems and Technology (TIST)},
  13(4):1--16, 2022.

\bibitem{kingma2013auto}
Diederik~P Kingma and Max Welling.
\newblock Auto-encoding variational bayes.
\newblock {\em arXiv preprint arXiv:1312.6114}, 2013.

\bibitem{lee2013pseudo}
Dong-Hyun Lee et~al.
\newblock Pseudo-label: The simple and efficient semi-supervised learning
  method for deep neural networks.
\newblock In {\em Workshop on challenges in representation learning, ICML},
  volume~3, page 896, 2013.

\bibitem{li2020federated}
Tian Li, Anit~Kumar Sahu, Ameet Talwalkar, and Virginia Smith.
\newblock Federated learning: Challenges, methods, and future directions.
\newblock {\em IEEE Signal Processing Magazine}, 37(3):50--60, 2020.

\bibitem{liu2021fate}
Yang Liu, Tao Fan, Tianjian Chen, Qian Xu, and Qiang Yang.
\newblock Fate: An industrial grade platform for collaborative learning with
  data protection.
\newblock {\em J. Mach. Learn. Res.}, 22(226):1--6, 2021.

\bibitem{liu2019communication}
Yang Liu, Yan Kang, Xinwei Zhang, Liping Li, Yong Cheng, Tianjian Chen, Mingyi
  Hong, and Qiang Yang.
\newblock A communication efficient collaborative learning framework for
  distributed features.
\newblock {\em arXiv preprint arXiv:1912.11187}, 2019.

\bibitem{liu2020federated}
Yang Liu, Yingting Liu, Zhijie Liu, Yuxuan Liang, Chuishi Meng, Junbo Zhang,
  and Yu Zheng.
\newblock Federated forest.
\newblock {\em IEEE Transactions on Big Data}, 2020.

\bibitem{liu2021defending}
Yang Liu, Zhihao Yi, Yan Kang, Yuanqin He, Wenhan Liu, Tianyuan Zou, and Qiang
  Yang.
\newblock Defending label inference and backdoor attacks in vertical federated
  learning.
\newblock {\em arXiv preprint arXiv:2112.05409}, 2021.

\bibitem{miyato2018virtual}
Takeru Miyato, Shin-ichi Maeda, Masanori Koyama, and Shin Ishii.
\newblock Virtual adversarial training: a regularization method for supervised
  and semi-supervised learning.
\newblock {\em IEEE transactions on pattern analysis and machine intelligence},
  41(8):1979--1993, 2018.

\bibitem{rasmus2015semi}
Antti Rasmus, Mathias Berglund, Mikko Honkala, Harri Valpola, and Tapani Raiko.
\newblock Semi-supervised learning with ladder networks.
\newblock {\em Advances in neural information processing systems}, 28, 2015.

\bibitem{romanini2021pyvertical}
Daniele Romanini, Adam~James Hall, Pavlos Papadopoulos, Tom Titcombe, Abbas
  Ismail, Tudor Cebere, Robert Sandmann, Robin Roehm, and Michael~A Hoeh.
\newblock Pyvertical: A vertical federated learning framework for multi-headed
  splitnn.
\newblock {\em arXiv preprint arXiv:2104.00489}, 2021.

\bibitem{sohn2020fixmatch}
Kihyuk Sohn, David Berthelot, Nicholas Carlini, Zizhao Zhang, Han Zhang,
  Colin~A Raffel, Ekin~Dogus Cubuk, Alexey Kurakin, and Chun-Liang Li.
\newblock Fixmatch: Simplifying semi-supervised learning with consistency and
  confidence.
\newblock {\em Advances in neural information processing systems}, 33:596--608,
  2020.

\bibitem{sun2022label}
Jiankai Sun, Xin Yang, Yuanshun Yao, and Chong Wang.
\newblock Label leakage and protection from forward embedding in vertical
  federated learning.
\newblock {\em arXiv preprint arXiv:2203.01451}, 2022.

\bibitem{vepakomma2018split}
Praneeth Vepakomma, Otkrist Gupta, Tristan Swedish, and Ramesh Raskar.
\newblock Split learning for health: Distributed deep learning without sharing
  raw patient data.
\newblock {\em arXiv preprint arXiv:1812.00564}, 2018.

\bibitem{wu2020privacy}
Yuncheng Wu, Shaofeng Cai, Xiaokui Xiao, Gang Chen, and Beng~Chin Ooi.
\newblock Privacy preserving vertical federated learning for tree-based models.
\newblock {\em arXiv preprint arXiv:2008.06170}, 2020.

\bibitem{wu2022practical}
Zhaomin Wu, Qinbin Li, and Bingsheng He.
\newblock Practical vertical federated learning with unsupervised
  representation learning.
\newblock {\em IEEE Transactions on Big Data}, 2022.

\bibitem{yang2021federated}
Dong Yang, Ziyue Xu, Wenqi Li, Andriy Myronenko, Holger~R Roth, Stephanie
  Harmon, Sheng Xu, Baris Turkbey, Evrim Turkbey, Xiaosong Wang, et~al.
\newblock Federated semi-supervised learning for covid region segmentation in
  chest ct using multi-national data from china, italy, japan.
\newblock {\em Medical image analysis}, 70:101992, 2021.

\bibitem{yang2019federated}
Qiang Yang, Yang Liu, Yong Cheng, Yan Kang, Tianjian Chen, and Han Yu.
\newblock {\em Federated Learning}.
\newblock Morgan \& Claypool Publishers, 2019.

\bibitem{yeh2009comparisons}
I-Cheng Yeh and Che-hui Lien.
\newblock The comparisons of data mining techniques for the predictive accuracy
  of probability of default of credit card clients.
\newblock {\em Expert systems with applications}, 36(2):2473--2480, 2009.

\bibitem{zhang2021secure}
Qingsong Zhang, Bin Gu, Cheng Deng, and Heng Huang.
\newblock Secure bilevel asynchronous vertical federated learning with backward
  updating.
\newblock In {\em Proceedings of the AAAI Conference on Artificial
  Intelligence}, volume~35, pages 10896--10904, 2021.

\bibitem{zhang2021improving}
Zhengming Zhang, Yaoqing Yang, Zhewei Yao, Yujun Yan, Joseph~E Gonzalez, Kannan
  Ramchandran, and Michael~W Mahoney.
\newblock Improving semi-supervised federated learning by reducing the gradient
  diversity of models.
\newblock In {\em 2021 IEEE International Conference on Big Data (Big Data)},
  pages 1214--1225. IEEE, 2021.

\bibitem{zhao2020semi}
Yuchen Zhao, Hanyang Liu, Honglin Li, Payam Barnaghi, and Hamed Haddadi.
\newblock Semi-supervised federated learning for activity recognition.
\newblock {\em arXiv preprint arXiv:2011.00851}, 2020.

\end{thebibliography}
}

\end{document}